\documentclass{article}
\usepackage{ijcai23}

\usepackage{times}
\usepackage{soul}
\usepackage{url}
\usepackage[hidelinks]{hyperref}
\usepackage[utf8]{inputenc}
\usepackage[small]{caption}
\usepackage{graphicx}
\usepackage{amsmath}
\usepackage{amsthm}
\usepackage{booktabs}
\usepackage{algorithm}
\usepackage{algorithmic}
\usepackage[switch]{lineno}

\usepackage{graphicx}
\usepackage{tikz}
\usepackage{forest}
\usetikzlibrary{trees,positioning,shapes,shadows,arrows.meta}
\usetikzlibrary{positioning} 
\usepackage{xcolor} 
\usepackage{multirow}
\usepackage{makecell}
\usepackage{amssymb}

\DeclareUnicodeCharacter{202F}{\,}

\linenumbers

\title{Acceleration Algorithms in GNNs: A Survey}

\author{
Lu Ma$^1$
\and
Zeang Sheng$^1$
\and
Xunkai Li$^3$
\and
Xinyi Gao$^4$
\and
Zhezheng Hao$^5$
\\
Ling Yang$^1$
\and
Wentao Zhang
$^2$
\And
Bin Cui$^{1}$
\affiliations
$^1$School of CS, Peking University, 
$^2$Center for machine learning research, Peking University\\
$^3$School of CS, Beijing Institute of Technology\\
$^4$The University of Queensland,
$^5$School of Artificial Intelligence, Northwestern Polytechnical University\\
\emails
\{cs.LuMa, haozhezheng\}@outlook.com,
\{shengzeang18, wentao.zhang, bin.cui\}@pku.edu.cn,
cs.xunkai.li@gmail.com,
xinyi.gao@uq.edu.au,
yangling0818@163.com
}
\begin{document}
\nolinenumbers
\maketitle
\begin{abstract}
Graph Neural Networks (GNNs) have demonstrated effectiveness in various graph-based tasks. However, their inefficiency in training and inference presents challenges for scaling up to real-world and large-scale graph applications. To address the critical challenges, a range of algorithms have been proposed to accelerate training and inference of GNNs, attracting increasing attention from the research community. In this paper, we present a systematic review of acceleration algorithms in GNNs, which can be categorized into three main topics based on their purpose: training acceleration, inference acceleration, and execution acceleration. Specifically, we summarize and categorize the existing approaches for each main topic, and provide detailed characterizations of the approaches within each category. Additionally, we review several libraries related to acceleration algorithms in GNNs and discuss our Scalable Graph Learning (SGL) library. Finally, we propose promising directions for future research. A complete summary is presented in our GitHub repository: \textcolor{blue}{\url{https://github.com/PKU-DAIR/SGL/blob/main/Awsome-GNN-Acceleration.md}}.
\end{abstract}

\section{Introduction}
Graph Neural Networks (GNNs) \cite{GCN,GAT}, which are able to extract graph structure information as well as interactions and potential connections between nodes, are currently the state-of-the-art in learning node and graph representations. GNNs demonstrates their effectiveness in various graph-based tasks, such as chemistry \cite{chemistry}, biology \cite{biology}, social network \cite{recommendersystems}, computer vision \cite{LSP}, and recommendation systems \cite{recommendersystems}. 

Despite the widespread use of GNNs, the training and inference of GNNs are known for their inefficiency, which presents significant challenges in scaling up GNNs for real-world and large-scale graph applications \cite{recommendersystems,pAScA}. Generally, GNNs rely on the adjacency matrix of the graph and the node features. However, the size of the graph data is growing exponentially. For instance, Facebook's social network graph contains over 2 billion nodes and 1 trillion edges. A graph of this scale can generate 100 terabytes of data during training. In addition, most GNNs are described using the message passing (MP) paradigm \cite{MPNN}, which is based on recursive neighborhood aggregation and transformation. The repetitive neighborhood expansion of GNNs leads to an expensive neighborhood expansion, resulting in significant computation and IO overhead during both training and inference.

{\bf Acceleration Algorithms in GNNs}, which combine training acceleration and inference acceleration, serve as a promising research direction to reduce the storage and computation consumption of GNNs. These algorithms have attracted considerable interests from the community. \cite{survey1} conduct a comprehensive survey on GNN acceleration methods from the perspective of algorithms, systems, customized hardware. \cite{survey2} review the acceleration algorithms for GNNs, focusing on graph-level and model-level optimizations. 
However, both of them have limitations in terms of comprehensive coverage and the absence of a taxonomy specifically focused on the purpose of algorithms. In contrast, we provide a systematic and comprehensive survey of acceleration algorithms in GNNs, focusing on three main topics: training acceleration, inference acceleration and execution acceleration (i.e., both training and inference). For training acceleration methods, we focus on graph sampling \cite{GraphSage} and GNN simplification \cite{SGC}. We focus on GNN knowledge distillation \cite{LSP}, GNN quantization \cite{Degree-Quant} and GNN pruning \cite{Channel-Pruning} as inference acceleration methods. For execution acceleration methods, we focus on GNN Binarization \cite{Bi-GCN} and graph condensation \cite{GCOND}. Then, we review some libraries related to acceleration in GNNs and discuss our Scalable Graph Learning (SGL) library. We hope that this survey aids in and advances the research and the applications of acceleration in GNNs.
\definecolor{mylightred}{RGB}{255,192,203}
\definecolor{mylightorange}{RGB}{255,218,185}
\tikzset{
  FARROW/.style={arrows={-{Latex[length=1.25mm, width=1.mm]}}, }, 
  U/.style = {circle, draw=melon!400, fill=melon, minimum width=1.4em, align=center, inner sep=0, outer sep=0},
  I/.style = {circle, draw=tea_green!400, fill=tea_green, minimum width=1.4em, align=center, inner sep=0, outer sep=0},
    cate0/.style = {rectangle, draw, minimum width=4cm, minimum height=1cm, align=left, rounded corners=3,draw=mylightred}, 
    cate1/.style = {rectangle, draw, minimum width=2cm, minimum height=0.8cm, align=left, rounded corners=3,draw=mylightred}, 
    cate2/.style = {rectangle, draw, text width=15cm, minimum height=0.8cm, align=left, rounded corners=3,draw=mylightred,fill=mylightorange}, 
    cate3/.style = {rectangle, draw, minimum width=3.7cm, minimum height=0.8cm, align=left, rounded corners=3,draw=mylightred},
  encoder/.style = {rectangle, fill=Madang!82, minimum width=10em, minimum height=3em, align=center, rounded corners=3},
}

\begin{figure*}
    \centering
    \resizebox{0.94\linewidth}{!}
    {
    \begin{tikzpicture}
    
    \node [cate0, distance=0cm, yshift=0cm, align=left,rotate=90,font=\large] (n1) at (0, 0) {\textbf{Acceleration Algorithms in GNNs}};
    
    \node [cate1, right of=n1, node distance=3cm, yshift=4cm] (n11) 
    {\textbf{GNN Training}};
    
    \node [cate1, below=4cm of n11.west, anchor=west] (n12) 
    {\textbf{GNN Inference}};
    
    \node [cate1, below=4cm of n12.west, node distance=5cm, anchor=west] (n13) {\textbf{GNN Execution}};

    \draw[] (n1.south) -- (n11.west);
    \draw[] (n1.south) -- (n12.west);
    \draw[] (n1.south) -- (n13.west);
    
    \node [cate3, right of=n11, node distance=4cm, yshift=3cm,] (n111) {\textbf{Graph Sampling}};

    \node [cate3, below=3.5cm of n111.west, anchor=west] (n112) {\textbf{GNN Simplification}};
    
    \draw[] (n11.east) -- (n111.west);
    \draw[] (n11.east) -- (n112.west);

    \node [cate2, right of=n111, node distance=11cm, yshift=2cm] (n1111) 
    {\textbf{Node-wise Sampling:} 
        GraphSage \cite{GraphSage},
        PinSage \cite{PinSage},
        VR-GCN  \cite{VR-GCN},
        BNS \cite{BNS},
        GNN-BS \cite{GNN-BS},
        PASS \cite{PASS}
        ANS-GT \cite{ANS-GT}
    };
    
    \node [cate2, below=1.3cm of n1111.west, anchor=west] (n1112) 
    {\textbf{Layer-wise Sampling:}
        FastGCN \cite{FastGCN},
        AS-GCN \cite{ASGCN},
        LADIES \cite{LADIES},
        GRAPES \cite{GRAPES}
    };

    \node [cate2, below=1.3cm of n1112.west, anchor=west] (n1113) 
    {\textbf{Graph-wise Sampling:}
        Cluster-GCN \cite{ClusterGCN},
        GraphSAINT \cite{GraphSAINT},
        LGCL \cite{LGCL},
        SHADOW-GNN  \cite{SHADOW-GNN},
        GNNAutoScale \cite{GNNAutoScale},
        MVS-GNN \cite{MVS-GNN},
        RWT\cite{RWT},
        LMC \cite{LMC}
    };

    \draw[] (n111.east) -- (n1111.west);
    \draw[] (n111.east) -- (n1112.west);
    \draw[] (n111.east) -- (n1113.west);
    
    \node [cate2, below=1.7cm of n1113.west, anchor=west] (n1121) 
    {\textbf{Simple model without attention:} 
        SGC \cite{SGC},
        S\textsuperscript{2}GC \cite{S2GC},
        SIGN \cite{SIGN},
        GBP \cite{GBP},
        PPRGo \cite{PPRGo},
        NDLS \cite{NDLS},
        AGP \cite{AGP},
        NAFS \cite{NAFS},
        APPNP \cite{APPNP},
        C\&S \cite{C&S}
    };
    
    \node [cate2, below=1.7cm of n1121.west, anchor=west] (n1122) 
    {\textbf{Difficult model with attention:}
        SAGN \cite{SAGN},
        GA-MLP \cite{GAMLP},
        PaSca \cite{pAScA},
        NIGCN \cite{NIGCN},
        SCARA \cite{SCARA},
        Grand+ \cite{GRAND+}
    };

    \draw[] (n112.east) -- (n1121.west);
    \draw[] (n112.east) -- (n1122.west);

    \node [cate3, below=2.55cm of n112.west, anchor=west] (n121) {\textbf{Knowledge Distillation}};
    
    \node [cate3, below=2cm of n121.west, anchor=west] (n122) {\textbf{GNN Quantization}};

    \node [cate3, below=1.3cm of n122.west, anchor=west] (n123) {\textbf{GNN Pruning}};
    
    \draw[] (n12.east) -- (n121.west);
    \draw[] (n12.east) -- (n122.west);
    \draw[] (n12.east) -- (n123.west);

    \node [cate2, below=1.3cm of n1122.west, anchor=west] (n1211) 
    {\textbf{GNN-to-GNN:} 
        LSP \cite{LSP},
        TinyGNN \cite{TinyGNN},
        G-CRD  \cite{G-CRD},
        GFKD \cite{GFKD},
        KDEP \cite{KDEP},
        GKD \cite{GKD}  
    };
    
    \node [cate2, below=1.3cm of n1211.west, anchor=west] (n1212) 
    {\textbf{GNN-to-MLP:}
        Graph-MLP \cite{Graph-MLP},
        GLNN \cite{GLNN},
        CPF \cite{CPF},
        NOSMOG \cite{NOSMOG},
        VQGraph \cite{VQGraph},
        KRD \cite{KRD},
        P\&D \cite{P&D}
    };

    \draw[] (n121.east) -- (n1211.west);
    \draw[] (n121.east) -- (n1212.west);

    \node [cate2, below=1.45cm of n1212.west, anchor=west] (n1221) 
    {LPGNAS \cite{LPGNAS},
     Degree-Quant \cite{Degree-Quant},
     SGQuant \cite{SGQuant},
     VQ-GNN \cite{VQ-GNN},
     A\textsuperscript{2}Q \cite{A2Q},
     EPQuant \cite{EPQuant},
     QLR \cite{QLR},
     \cite{Haar}
    };
    \draw[] (n122.east) -- (n1221.west);

    \node [cate2, below=1.3cm of n1221.west, anchor=west] (n1231) 
    {\cite{Channel-Pruning},
     UGS \cite{UGS},
     ICPG\cite{Inductive},
     \cite{GEB},
     DLTH \cite{DLTH},
     \cite{improved-UGS},
     Snowflake \cite{Snowflake}
    };
    \draw[] (n123.east) -- (n1231.west);

    \node [cate3, below=1.2cm of n123.west, anchor=west] (n131) 
    {\textbf{GNN Binarization}};
    
    \node [cate3, below=1.4cm of n131.west, anchor=west] (n132) 
    {\textbf{Graph Condensation}};
    
    \draw[] (n13.east) -- (n131.west);
    \draw[] (n13.east) -- (n132.west);

    \node [cate2, below=1.2cm of n1231.west, anchor=west] (n1311) 
    {Bi-GCN \cite{Bi-GCN},
     BGN \cite{BGN},
     \cite{BinaryGNN},
     Meta-Aggregator \cite{Meta-Aggregator},
     BitGNN \cite{BitGNN}
    };
    \draw[] (n131.east) -- (n1311.west);

    \node [cate2, below=1.1cm of n1311.west, anchor=west] (n1321) 
    {
     \textbf{Node classification:} 
     GCond \cite{GCOND},
     SFGC \cite{SFGC},
     MCond \cite{MCond},
     GCDM \cite{GCDM},
     GCEM \cite{GCEM},
     GC-SNTK \cite{GC-SNTK}
    };
    \draw[] (n132.east) -- (n1321.west);

    \node [cate2, below=1cm of n1321.west, anchor=west] (n1321) 
    {
     \textbf{Graph classification:}
     DosCond \cite{DosCond},
     KIDD \cite{KIDD}
    };
    \draw[] (n132.east) -- (n1321.west);

    \end{tikzpicture}
    
    }
    \caption
    {
        {Taxonomy of GNN acceleration algorithms. We comprehensively introduce GNNs training acceleration in Section 3, GNNs inference acceleration in Section 4 and GNNs execution acceleration in Section 5, followed by our discussions of related libraries in Section 6 and future research directions in Section 7.}
    }
    \label{fig:user-item-summ-fig}
\end{figure*}
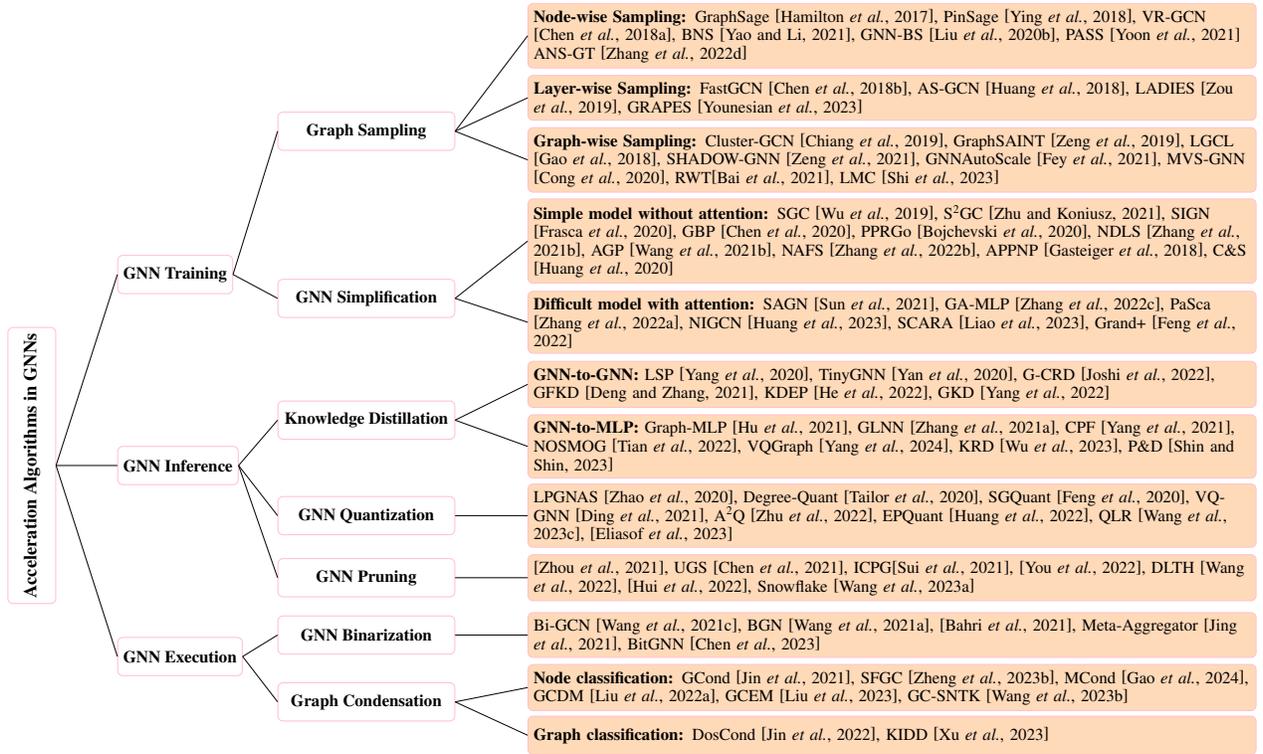

\section{Preliminary}

In this section, we first provide a brief introduction to the fundamental formulations and paradigm of GNNs. Subsequently, we discuss two aspects of challenges associated with acceleration algorithms in GNNs: the time complexity challenges and the memory complexity challenges, before moving on to the next section.

\subsection{Graph Neural Networks}
Consider a graph $\mathcal{G}=(\mathcal{V}, \mathcal{E})$, with nodes $\mathcal{V}$ and edges $\mathcal{E}$. Node features are represented as $\mathbf{X} = \{\mathbf{x}_1,\dots,\mathbf{x}_{n-1},\mathbf{x}_{n} \} \in \mathbb{R}^{n \times f}$, where $n$ is the number of nodes, and $f$ is the dimension of node features.  
The neighborhood set and the degree of node $v$ are denoted by $\mathcal{N}(v)$ and $d(v)$, respectively.
The adjacency matrix is defined as $\mathbf{A} \in \left\{ 0,1 \right\}^{n\times n}$. 
$\mathbf{A}\left[u,v\right] = 1$ if $(u,v) \in \mathcal{E}$, otherwise, $\mathbf{A}\left[u,v\right] = 0$.
We use bold uppercases (e.g., $\mathbf{X}$) and bold lowercases (e.g., $\mathbf{x}$) to represent matrices and vectors, respectively.

Most GNNs can be formulated using the MP paradigm (e.g., GCN \cite{GCN}, GraphSAGE \cite{GraphSage}, GAT \cite{GAT}, etc.). In this paradigm, each layer of the GNNs adopts an aggregation function and an update function. The $l^{th}$ MP layer of GNNs is formulated as follows:
\begin{equation}
\begin{aligned}
    \mathbf{m}_v^{(l)} \leftarrow \operatorname{aggre}&\operatorname{gate}  
    \left( \mathbf{h}_v^{(l-1)},\left\{\mathbf{h}_u^{(l-1)} \mid u \in \mathcal{N}(v)\right\}\right), \\
    \mathbf{h}_v^{(l)} &\leftarrow \operatorname{update} 
    \left( \mathbf{m}_v^{(l)},\mathbf{h}_v^{(l-1)} \right),
\end{aligned}
\end{equation}
where $\mathbf{h}_v^{(l)}$ represents the representation of node $v$ in the $l^{th}$ layer, while $\mathbf{m}_v^{(l)}$ denotes the message for node $v$. 
The aggregation function is represented by $\operatorname{aggregate}$, and the update function is denoted by $\operatorname{update}$.
A message vector $\mathbf{m}_v^{(l)}$ is computed using the representations of its neighboring nodes $\mathcal{N}(v)$ through an aggregation function, and $\mathbf{m}_v^{(l)}$ is subsequently updated by an update function. The node representation is typically initialized as node features, denoted by $\mathbf{H}^{(0)} = \mathbf{X}$, and the final representation is obtained after $L$ MP layers as $\mathbf{H} = \mathbf{H}^{(L)}$.

The Graph Convolutional Network (GCN) \cite{GCN}, one of the earliest GNNs, operates by performing a linear approximation to spectral graph convolutions. The $l^{th}$ MP layer is formulated using the MP paradigm as follows:
\begin{gather*}
    \mathbf{m}_v^{(l)} = \sum_{u \in \mathcal{N}(v)}  \mathbf{h}_u^{(l-1)} / \sqrt{\tilde{d}_v \tilde{d}_u},
    \\
    \mathbf{h}_v^{(l)} = \sigma\left(\mathbf{W}^{(l)}\mathbf{m}_v^{(l)}\right),
\end{gather*}
where $\tilde{d}(v)$ is the degree of node $v$ obtained from the adjacency matrix with self-connections, denoted by $\tilde{\mathbf{A}}=\mathbf{I}+\mathbf{A}$; and $\mathbf{W}^{(l)}$ signifies learnable weights.

\subsection{Challenges of scalable GNNs}
GNNs are notorious for inefficient training and inference. The challenges are twofold: time complexity challenges and memory complexity challenges.
\\
{\bf Time Complexity Challenges:}
 Scalable GNNs frequently encounter challenges related to time complexity during both training and inference phases. 
 Given that the nodes in a graph cannot be treated as independent samples, most GNNs necessitate the resource-intensive and time-consuming process of Message Passing (MP) during both training and inference.
 The computation of the hidden representation for a specific node involves the aggregation of information from its neighbors. Consequently, the neighbors must also consider their own neighbors, resulting in an exponential expansion with each layer.
 This exponential growth equates to substantial computational costs. Simultaneously, due to the irregularity of graphs, data is often not stored contiguously in memory, leading to high I/O overhead during aggregation. The processes of aggregating information, performing transformations, and updating node representations pose time complexity challenges.
\\
{\bf Memory Complexity Challenges:}
Another significant challenge of scalable GNNs pertains to memory complexity. Training full-batch GNNs, such as GCN, necessitates storing the entire graph in GPU memory, leading to substantial memory overflow issues, especially when dealing with graphs containing millions of nodes. Even for batch-training methods, considerable memory resources are required to process large-scale graphs. Storing and accessing graph data in a more memory-efficient manner presents challenges. Moreover, it is essential to store the activation output of each layer to compute the gradient during the backward pass. As the number of GNN layers increases, the activation footprints progressively occupy more GPU memory during the training step.

Approaches with acceleration in training, inference, and execution, reviewed in later sections, aim to address the two aspects of challenges.

\section{Training Acceleration}
In this section, we discuss two types of training acceleration methods, graph sampling and GNN simplification. For each category of methods, we elucidate the fundamental definition and properties, followed by exemplification of typical work that falls within these categories

\subsection{Graph Sampling}
Graph sampling, a common yet sophisticated technique, is employed to accelerate the training of GNNs. This technique performs batch-training by utilizing sampled subgraphs to approximate node representations. It can mitigate the issue of neighbor explosion, thereby significantly reducing memory consumption during training. For the sake of completeness, we reiterate the unified formulation of graph sampling methods as follows:
\begin{equation}
    \begin{aligned}
        \mathbf{m}_v^{(l)} \leftarrow \operatorname{aggre}&\operatorname{gate} 
        \left( \mathbf{h}_v^{(l-1)}, \left\{\mathbf{h}_u^{(l-1)} \mid (u,v) \in Block^{(l)} \right\}\right),
        \\
        &\mathbf{h}_v^{(l)} \leftarrow \operatorname{update} \left( \mathbf{m}_v^{(l)}, \mathbf{h}_v^{(l-1)} \right),
    \end{aligned}
\end{equation}%
where $Block^{(l)} = \{(u,v) \mid u \in n^{(l-1)}, v \in n^{(l)} ,(u,v) \in \mathcal{E}\}$, and $n^{(l)}$ represents the set of sampled nodes at the $l^{th}$ MP layer. Formally, the key distinction among graph sampling methods lies in how the sets $\{n^{(0)},\dots, n^{(L-1)}, n^{(L)} \}$ are sampled. Based on the scale at which these methods sample $n^{(l)}$, we can classify them into node-wise sampling, layer-wise sampling, and graph-wise sampling methods. 
\\
{\bf Node-wise Sampling:} $n^{(l)} = \bigcup_{v \in n^{(l+1)}} \{ u \mid u \sim r \cdot \mathbb{P}_{\mathcal{N}(v)}\}$, where $\mathbb{P}$ represents a sampling distribution; $\mathcal{N}(v)$ is the sampling space; and $r$ denotes the number of samples. Node-wise sampling methods focus on sampling a fixed-size set of neighbors for each node in each MP layer, which mitigates the issue of neighbor explosion and reduces the memory complexity from $\mathcal{O}(d^L)$ to $\mathcal{O}(r^L)$, where $d$ is the average node degree. GraphSAGE \cite{GraphSage} is the pioneering work in graph sampling, which selects a fixed-size set of target nodes as a mini-batch and randomly samples a tree rooted at each target node by recursively expanding the root node's neighborhood. For each sampled tree, GraphSAGE computes the hidden representation of the target node by aggregating the node representations from bottom to top. Furthermore, PinSAGE \cite{PinSage} builds upon GraphSAGE by incorporating an importance score to each neighbor through random walks as sampling probability. VR-GCN \cite{VR-GCN} utilizes historical representations from the previous epoch to regulate the variance from neighbor sampling. However, VR-GCN needs to store the previous representations of all nodes and suffers from increased complexity. GNN-BS \cite{GNN-BS} formulates the neighbor sampling as a bandit problem, and proposes a learnable sampler aimed at minimal variance. 
\\
{\bf Layer-wise sampling:} $n^{(l)} =  \{ u \mid u \sim r \cdot \mathbb{P}_{\mathcal{V}} \}$. Layer-wise sampling methods focus on sampling a fixed-size set of nodes for each MP layer, which reduces the memory complexity from $\mathcal{O}(r^L)$ to $\mathcal{O}(rL)$. As the inaugural layer-wise sampling method, FastGCN \cite{FastGCN} samples a fixed number of nodes independently in each layer by the node-degree-based probability distribution: $p(u) \propto d(u)$. LADIES \cite{LADIES} employs the same iid as FastGCN but restricts the sampling domain to the neighborhood of the sampled nodes. AS-GCN \cite{ASGCN} enhances FastGCN by introducing an adaptive sampler for explicit variance reduction, ensuring higher accuracy. Although layer-wise sampling methods successfully control the neighbor expansion, the computation graphs may become too sparse to maintain high accuracy.
\\
{\bf Graph-wise sampling:} $n^{(0)}=\dots= n^{(L-1)}= n^{(L)} = \{ u \mid u \sim r \cdot \mathbb{P}_{\mathcal{G}} \}$. Graph-wise sampling methods concentrate on sampling the same sub-graph for each MP layer based on a specific sampling strategy $\mathbb{P}_{\mathcal{G}}$. The sampled sub-graph is significantly smaller, thus enabling us to conduct full-batch MP on it without concern for memory overflow issues. ClusterGCN \cite{ClusterGCN} first partitions the entire graph into smaller non-overlapping clusters using clustering algorithms (i.e., METIS \cite{METIS}). During each training iteration, ClusterGCN selects certain clusters as a sub-graph for further full-batch training. GraphSAINT \cite{GraphSAINT} adopts an enhanced approach that corrects for the bias and variance of the estimators when sampling subgraphs for training, which can achieve high accuracy and rapid convergence. SHADOW-GNN \cite{SHADOW-GNN} proposes to extract subgraphs for each target node and then apply GNNs on the subgraphs to address the scalability challenge. Moreover, GNNAutoScale \cite{GNNAutoScale} uses historical embeddings to generate messages outside the sampled sub-graph, thereby maintaining the expressiveness of the original GNNs. Graph-wise sampling methods are applicable to a broad range of GNNs by directly running them on subgraphs, but the partitioning of the original graph could have a significant impact on the stability of training

\begin{table*}[ht]
\caption{Summary of time complexity and memory complexity for GNN training acceleration methods. $n$, $m$, $c$, and $f$ are the number of nodes, edges, classes, and feature dimensions, respectively. $b$ is the batch size, and $r$ refers to the number of sampled nodes. $L$ and $K$ corresponds to the number of times we aggregate features and labels, respectively. Besides, $P$ and $Q$ are the number of layers in MLP classifiers trained with features and labels, respectively. For GraphSAINT, $d = \frac{m}{n}$ is the average degree of $\mathcal{G}$ and the number of batches is $\frac{n}{b}$. For GBP, $\epsilon$ denote the error threshold. For simplicity we omit the memory for storing the graph or sub-graphs since they are fixed and usually not the main bottleneck.}
\small
\centering
\begin{tabular}{cc|l|ll}
\hline  \textbf{Type} &  \textbf{Model} & 
    \begin{tabular}{l} 
    \textbf{Pre-processing} \\ \textbf{Time}
    \end{tabular} & 
    \begin{tabular}{l} 
    \textbf{Training} \\ \textbf{Time}
    \end{tabular} & 
    \begin{tabular}{l} 
    \textbf{Training} \\ \textbf{Memory}    
    \end{tabular}\\
\hline 
& GraphSAGE & - & $\mathcal{O}\left(k^L n f^2\right)$ & $\mathcal{O}\left(b r^L f+L f^2\right)$ \\
&VR-GCN & - & $\mathcal{O}\left(L m f+L n f^2+r^Lnf^2\right)$ & $\mathcal{O}\left(L n f+L f^2\right)$ \\
Graph sampling & FastGCN & - & $\mathcal{O}\left(k L n f^2\right)$  & $\mathcal{O}\left(b k L f+L f^2\right)$ \\
& Cluster-GCN & $\mathcal{O}(m)$ & $\mathcal{O}\left(L m f+L n f^2\right)$  & $\mathcal{O}\left(b L f+L f^2\right)$ \\
&GraphSAINT & - & $\mathcal{O}\left(L bd f+L n f^2\right)$ & $\mathcal{O}\left(bLf+Lf^2\right)$\\
\hline & SGC & $\mathcal{O}(L m f)$ & $\mathcal{O}\left(n f^2\right)$  & $\mathcal{O}\left(b f+f^2\right)$ \\
& S\textsuperscript{2}GC  & $\mathcal{O}(L m f)$ & $\mathcal{O}\left(n f^2\right)$  & $\mathcal{O}\left(b f+f^2\right)$ \\
GNN simplification& SIGN & $\mathcal{O}(L m f)$ & $\mathcal{O}\left(L P n f^2\right)$ & $\mathcal{O}\left(bLf + P f^2\right)$ \\
without attention& GBP & $\mathcal{O}\left(L n f+L \frac{\sqrt{m \lg n}}{\varepsilon}\right)$ & $\mathcal{O}\left(P n f^2\right)$ & $\mathcal{O}\left(b f+P f^2\right)$ \\
& APPNP & - & $\mathcal{O}\left(L m f + P n f^2\right)$ & $\mathcal{O}\left(n f+P f^2\right)$ \\
& C\&S & $\mathcal{O}\left(K m c\right)$ & $\mathcal{O}\left(P n f^2\right)$ & $\mathcal{O}\left(b f+P f^2\right)$ \\
\hline GNN simplification&SAGN & $\mathcal{O}(L m f)$ & $\mathcal{O}\left(L P n f^2\right)$ & $\mathcal{O}\left(bLf + P f^2\right)$ \\
with attention &GAMLP & $\mathcal{O}(L m f + K m c)$ & $\mathcal{O}\left(P n f^2 +  Q n c^2\right)$ & $\mathcal{O}\left(bf + P f^2 + Q c^2\right)$ \\
\hline
\end{tabular}
\label{tab:booktabs}
\end{table*}

\subsection{GNN Simplification}
\begin{figure}
    \centering
    \includegraphics[width=0.45\textwidth]{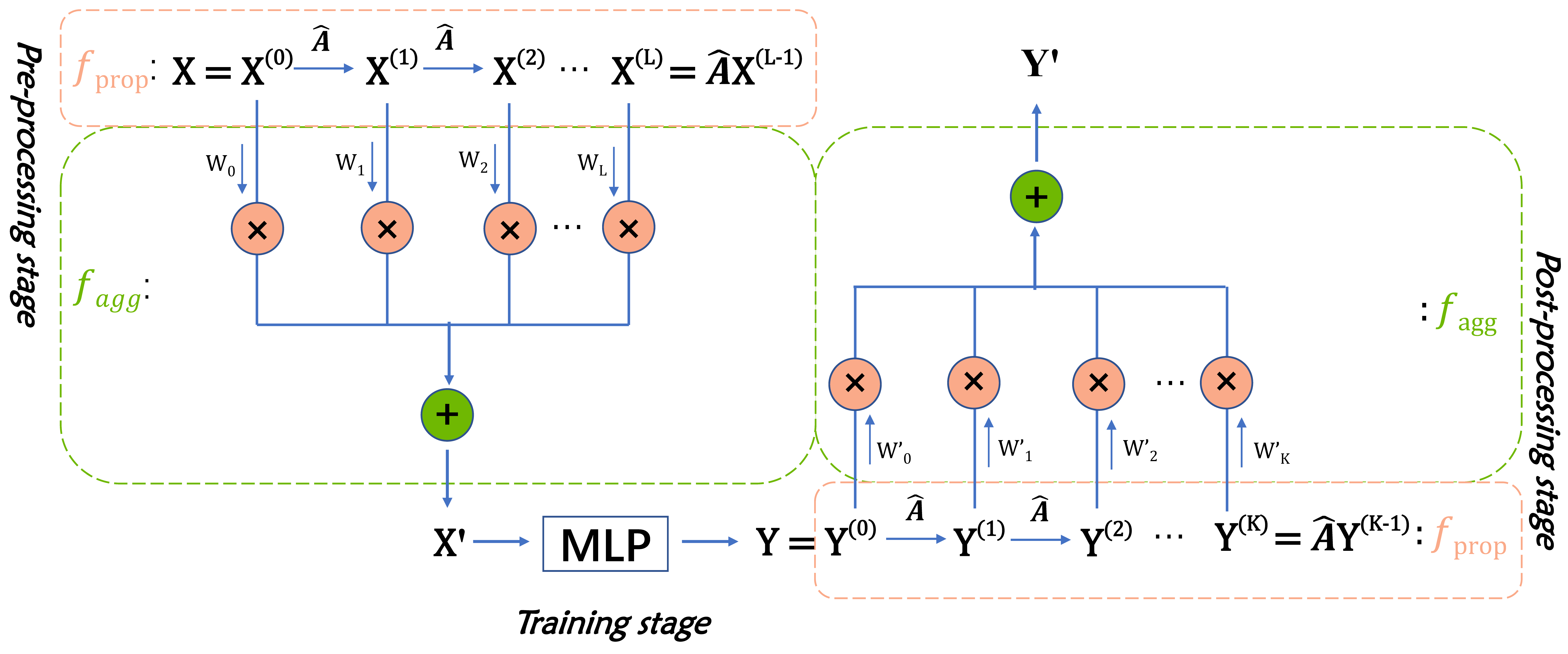}
    \caption{  
        Illustration of SGAP for GNN simplification methods.
    }
    \label{fig:SGAP}
\end{figure}

GNN simplification is a recent direction to accelerate GNN training, which involves decoupling the standard MP paradigm. The primary concept of GNN simplification methods is to either precompute the propagated features during pre-processing stages or propagate predictions during post-processing stages, thereby separating the propagation operation from the transformation operation. These GNN simplification methods offer two primary advantages. On one hand, these methods can directly utilize sparse matrix multiplication, conducted only once at CPUs, to obtain propagated features or propagated predictions, rather than relying on an inefficient MP process, which significantly reduces time complexity. On the other hand, because the dependencies between nodes have been fully addressed in the pre- or post-processing stage, it becomes feasible to divide the training nodes into smaller mini-batches. This enables batch-training, thereby significantly reducing memory consumption during training. However, GNN simplification models are not sufficiently expressive due to the absence of a trainable aggregation process, which limits their applications. To reflect the development of these methods, we categorize them into two types: simple methods without attention techniques and complex methods with attention techniques.
\\
{\bf Simple Model without Attention:} SGC \cite{SGC}, which initially highlights that empirically, the removal of intermediate nonlinearities does not affect model performance, is the pioneering work in GNN simplification. The formulation of SGC is as follows:
\begin{equation}
\begin{split}
    Y &= \operatorname{softmax} \left( \tilde{\mathbf{A}}\cdots \tilde{\mathbf{A}}\mathbf{X}\mathbf{W}^{(1)}\cdots\mathbf{W}^{(L)} \right) 
    \\
    &= \operatorname{softmax} \left( \tilde{\mathbf{A}}^L\mathbf{X}\mathbf{W} \right),
\end{split}
\end{equation}
where the propagated features $\tilde{\mathbf{A}}^L$ can be precomputed, resulting in a significant reduction in training time. Following the design principle of SGC, \cite{SIGN} propose SIGN, which employs various local graph operators to concatenate the propagated features of different iterations, thereby achieving superior performance compared to SGC. \cite{S2GC} introduce S\textsuperscript{2}GC, which averages the propagated features across different iterations as $\sum_{l=0}^{L}\mathbf{\tilde{A}}^l\mathbf{X}$. GBP \cite{GBP} further refines the combination process through weighted averaging as $\mathbf{P} = \sum_{l=0}^{L}w_l\tilde{\mathbf{A}}^l\mathbf{X}$ with $w_l = \beta\left(1-\beta\right)^l$ and proposes an approximation algorithm to compute $\mathbf{P}$. AGP \cite{AGP} proposes a unified randomized algorithm capable of computing various proximity queries and facilitating feature propagation. To better utilize the propagated features across different iterations, NDLS \cite{NDLS} theoretically analyzes what influences the propagation iterations and provides a bound to guide how to control the iterations for different nodes. Meanwhile, NAFS \cite{NAFS} uses the node-adaptive weight to average the different iterations of propagated features and proposes different ensemble strategies to obtain the final node embeddings. In addition to the aforementioned precomputating methods, there are some label propagation methods which can accelerate training. PPNP \cite{APPNP} trains on raw features, and then propagates predictions using a personalized PageRank matrix \cite{PageRank} in post-processing process. Directly calculating the personalized PageRank matrix is inefficient and a fast approximation of PPNP, called APPNP, is introduced. However, APPNP executes propagation operations in each training epoch, which makes it only faster than GCN and challenging to perform on large-scale graphs. Building upon APPNP, C\&S \cite{C&S} utilizes label propagation not only for predictions smoothness but also for error correction, which is faster to train and easily scales to large-scale graphs.
\\
{\bf Complex Model with Attention:} To enhance the performance of the aforementioned simple methods, SAGN \cite{SAGN} replaces the redundant concatenation operation in SIGN with attention techniques. This technique effectively collects neighbor information from various hops in an adaptive manner, which improve the prediction performance. Moreover, GAMLP \cite{GAMLP} defines three well-defined attention techniques, and each node in GAMLP has the capability to utilize the propagated features across different iterations. Furthermore, PaSca \cite{pAScA} provides a unified framework of GNN simplification methods and introduces the Scalable Graph Neural Architecture Paradigm (SGAP) for GNN simplification methods. As depicted in Figure \ref{fig:SGAP}, SGAP typically involves three independent stages: (1) Pre-processing: propagates raw features to generate propagated features for each iteration and aggregates these propagated features to generate the final combined message for each node, (2) Training: feeds the propagated and aggregated information into a multi-layer perceptron (MLP) for training, and (3) Post-processing: propagates predictions to generate propagated predictions and then aggregates them to obtain the final predictions. The formulation of SGAP is as follows:
\begin{equation}
\footnotesize
\begin{aligned}
    \operatorname{Pre-processing: }&\quad \mathbf{M} \leftarrow f_{\mathrm{prop}}\left(\tilde{\mathbf{A}},\mathbf{X}\right) 
    ,
    \mathbf{X}^{'} \leftarrow f_{\mathrm{agg}}\left(\mathbf{M}\right)
    ,
    \\
    \operatorname{Training: }&\quad \mathbf{Y} = \operatorname{MLP}\left(\mathbf{X}^{'}\right),
    \\
    \operatorname{Post-processing: }
    &\quad \mathbf{M}^{'} \leftarrow f_{\mathrm{prop}}\left(\tilde{\mathbf{A}},\mathbf{Y}\right) 
    ,
    \mathbf{Y}^{'} \leftarrow f_{\mathrm{agg}}\left(\mathbf{M}^{'}\right),
\end{aligned}
\end{equation}
where $\mathbf{M} = \{\mathbf{X}^{(0)},\mathbf{X}^{(1)},\dots,\mathbf{X}^{(L)}\}$ represents the set of propagated features, with $\mathbf{X}^{(i+1)}=\tilde{\mathbf{A}}\mathbf{X}^{(i)}$, and $\mathbf{X}^{(0)} = \mathbf{X}$; $\mathbf{M}^{'} = \{\mathbf{Y}^{(0)},\mathbf{Y}^{(1)},\dots,\mathbf{Y}^{(K)}\}$ denotes the set of propagated predictions, with $\mathbf{Y}^{(i+1)}=\tilde{\mathbf{A}}\mathbf{Y}^{(i)}$, and $\mathbf{Y}^{(0)} = \mathbf{Y}$; $f_{\mathrm{prop}}$ signifies propagation function, and $f_{\mathrm{agg}}$ denotes aggregation function. Generally, SGAP first obtains propagated features or predictions across different iterations using various propagation matrices. Subsequently, SGAP aggregates these propagated features or predictions, either with or without attention techniques.

Building upon on the SGAP paradigm, PaSca introduces a comprehensive design space, encompassing 150k possible designs of GNN simplification methods(e.g., SGC, SIGN, GAMLP, etc.). Rather than focusing on individual designs, PaSca prioritizes the overall design space, leading to superior performance and scalability compared to the STOA.

\section{GNN Inference Acceleration}
Despite the great performance of GNNs on a lot of tasks, due to their inefficient inference, their inefficient inference makes them less favored when deployed in latency-sensitive applications \cite{TinyGNN}. In this section, we discuss three types of inference acceleration methods: knowledge distillation (KD), GNN quantization, and GNN pruning. For each category, we fisrt introduce the fundamental definition and properties, followed by examples of typical works.

\subsection{Knowledge Distillation}
KD is a learning paradigm that extracts knowledge from a teacher model and transfers it to a student model \cite{KD}. Its objective is to enable the student model to mimic the teacher model through knowledge, where the logits, activations, and features can all be considered as knowledge. During training, KD guides the student model using the KD loss $\mathcal{L}_{KD}$, which is typically computed by a divergence function (e.g., KullbackLeibler divergence $D_{KL}$) between the knowledge of the student $k^{s}$ and the teacher $k^{t}$. This process can be formulated as follows:
\begin{equation}
    \mathcal{L}_{KD} = D_{KL}(k^{t},k^{s}).
\end{equation}

KD on GNNs aims either at performance improvement or at model compression, which also implies inference acceleration \cite{KDG}. In the case of the latter, the student model is deemed more efficient than the teacher model, with either fewer parameters or a different model structure offering better scalability(e.g. MLP). KD on GNNs, aimed at inference acceleraion, can be classified into two categories: GNN-to-GNN KD methods and GNN-to-MLP KD methods, depending on the type of their student models. 
\\
{\bf GNN-to-GNN KD:} LSP \cite{LSP} preserves and distills local structures of nodes as knowledge from a teacher GNN to a student GNN. LSP models the local structure of node $v$ via vectors that calculate the similarity between $v$ and its one-hop neighbors as:
\begin{equation}
    LS_v(u) = \frac{e^{\operatorname{SIM}(h_v,h_u)}}{\sum_{i \in \mathcal{N}(v)}{ e^{\operatorname{SIM}(h_v,h_i)} }}, \space \space  u \in \mathcal{N}(v),
\end{equation}
where $\operatorname{SIM}(h_i, h_j) = \Vert h_i - h_j \Vert_2^2$. LSP demonstrates excellent distillation performance across multiple domains and paves the way for KD on GNNs. TinyGNN \cite{TinyGNN} leverages peer node information and adopts a neighbor distillation strategy to implicitly acquire local structure knowledge by the Peer-Aware Module, which results in a significant speed-up in student GNN inference. GFKD \cite{GFKD} transfers knowledge from a teacher GNN to a student GNN by generating fake graphs. To further improve the performance, G-CRD \cite{G-CRD} not only preserves local structural information but also implicitly preserves the global topology information through contrastive learning, and GKD \cite{GKD} leverages a neural heat kernel to capture the global structure.
\\
{\bf GNN-to-MLP KD:} GNN-to-GNN KD methods apply KD to train student GNNs with fewer parameters, thereby accelerating GNN inference. However, the time-consuming MP is still required during inference. To address the time complexity issue, recent studies have explored distilling knowledge from trained GNNs into MLPs, which do not rely on graph data and can be efficiently deployed in latency-sensitive applications without the time-consuming MP. 
Graph-MLP \cite{Graph-MLP} attempts to train an MLP student model using a neighbor contrastive loss. GLNN \cite{GLNN} proposes training an MLP model with additional soft label prediction of a teacher GNN as targets to distill the topology knowledge. NOSMOG \cite{NOSMOG} incorporates position features to inject structural information into MLPs and utilizes adversarial feature augmentation to enhance the robustness of MLPs. Moreover, KRD \cite{KRD} explores the reliability of different nodes in a teacher GNN and proposes to sample a set of additional reliable knowledge nodes as supervision for training an MLP. VQGraph \cite{VQGraph} utilizes a structure-aware codebook constructed by vector-quantized variational autoencoder (VQ-VAE) to sufficiently transfer the structural knowledge from GNN to MLP.

\subsection{GNN Quantization}
Model quantization refers to the process of mapping continuous data (e.g., parameters, weights, and activations of neural networks) to smaller-sized representations. This process aims to reduce both the computational and memory consumption of models \cite{Quantization}. It primarily involves converting high-precision numerical values into lower precision representations (e.g., from 32-bit floating point numbers to 8-bit integers). Quantization can be categorized into two types: Quantization Aware Training (QAT), which involves training a model with an awareness of the quantization process, and Post-Training Quantization (PTQ), which applies quantization to a pre-trained model after training. As for GNNs, most quantization methods focus on enabling the usage of low-precision integer arithmetic during inference, which facilitates quicker model inference and reduces memory usage.

\cite{Degree-Quant} analyzes the reasons for the failure of QAT for GNNs and propose a QAT method for GNNs, termed Degree-Quant. This method involves the selection of high in-degree nodes for full-precision training, while all other nodes are quantized. Despite the known degradation of model performance by QAT, quantized GNNs not only perform comparably to FP32 models in most cases, but also achieve up to 4.7× speedups compared to FP32 models, with 4-8× reductions in memory usage. \cite{A2Q} proposes the Aggregation-Aware mixed-precision Quantization (A\textsuperscript{2}Q) for GNNs, where an optimal bit-width is automatically learned and assigned to each node in the graph. A\textsuperscript{2}Q can achieve up to an 18.6x compression ratio with negligible accuracy degradation, thereby reducing the time and memory consumption of GNNs. In addition to the aforementioned QAT methods, LPGNAS \cite{LPGNAS} quantizes GNNs using Network Architecture Search (NAS). VQ-GNN \cite{VQ-GNN} employs vector quantization (VQ) to reduce the dimensions of features and parameters of GNNs. SGQuant \cite{SGQuant} introduces a PTQ method for GNNs, which quantizes the node features while keeping the weights at full precision, thus limiting its efficiency and usage.

\subsection{GNN Pruning}
The technique aimed at reducing unnecessary parameters in neural networks is referred to as pruning \cite{Pruning}. Pruning is commonly employed to reduce the size or computational consumption of the trained networks, accelerating inference without compromising accuracy. Pruning provides a trade-off between inference speed and accuracy. As more parameters are removed, the model is likely to infer faster but with less accuracy, and vice versa.

To accelerate large-scale and real-time GNN inference, \cite{Channel-Pruning} presents a method for pruning the dimensions in each GNN layer, which can be applied to most GNN architectures with minimal or no loss of accuracy. \cite{UGS} initially extends the lottery ticket hypothesis \cite{LotteryTicket} to GNNs, introducing the Graph Lottery Ticket (GLT) hypothesis and a unified GNN sparsification (UGS) framework. This framework can simultaneously prune the graph adjacency matrix and the model weights, resulting in faster GNN inference. Building upon GLT hypothesis, \cite{DLTH} presents the Dual Graph Lottery Ticket framework to transform a random ticket into a triple-win graph lottery ticket with high sparsity, high performance, and good explainability. Meanwhile, \cite{improved-UGS} proposes two special techniques to improve the performance UGS in high graph sparsity scenarios. 

\section{GNN Execution Acceleration}
GNN execution acceleration methods can accelerate both training and inference of GNNs. In this section, we discuss two types of GNN execution acceleration methods, called GNN Binarization and Graph Condensation. We first discuss the fundamental definition and property of these methods, and exemplify typical work belonging to these categories.

\subsection{GNN Binarization}
Binarization is a technique that pushes model quantization to the extreme by using only a single bit for weights and activations \cite{Binarization}. Binarization can significantly reduce the storage needs and computational operations required for both training and inference. Binarization methods have achieved great success in compressing models and providing significant acceleration in both inference and training.

Bi-GCN \cite{Bi-GCN} introduces binarization techniques to GNNs, which binarizes both the network parameters and node features with minimal accuracy loss, leading to faster inference. Bi-GCN also proposes a novel binary gradient approximation-based back-propagation technique for effectively training binary GCNs. \cite{BGN} proposes BGN with binarized vectors and bit-wise operations, which can not only reduce memory consumption but also accelerate inference. In addition, the gradient estimator allows BGN to efficiently perform back-propagation to accelerate the training process. Meta-Aggregator \cite{Meta-Aggregator} introduces two aggregators: the Greedy Gumbel Aggregator (GNA) and Adaptable Hybrid Aggregator (ANA), to enhance the binary training accuracy during the aggregation phase.

\subsection{Graph Condensation}
Graph condensation is a technique aimed at minimizing the performance gap between GNN models trained on a synthetic, simplified graph and the original training graph \cite{GCOND}. Graph condensation can condense a large, original graph into a smaller, synthetic graph while retaining high levels of information. After obtaining synthetic and highly informative graphs, these graphs can be utilized to train and infer the second and subsequent GNNs, achieving performance comparable to the GNN trained on the original graphs. Graph condensation leads to acceleration in both training and inference, which can be utilized for node classification tasks or graph classification tasks.
\\
{\bf Node classification:} \cite{GCOND} first introduces GCond for graph condensation, which matches the gradients between GNNs trained on the original large-scale graph and the synthetic graph. GCond is shown to maintain over 95\% original test accuracy while reducing their graph size by more than 99.9\%, and the same condensed graph can be transferred to train different GNNs to achieve good performance. Moreover, SFGC \cite{SFGC} proposes to distill large-scale graphs to small-scale synthetic graph-free data to compress the structural information into the node features, and GCDM \cite{GCDM} proposes distribution matching for graph condensation. Additionally, MCond \cite{MCond} focuses more on accelerating inference of GNNs, explicitly learning a sparse mapping matrix between original and synthetic nodes. To perform inference on inductive nodes, MCond seamlessly integrates new nodes into the synthetic graph, which is both efficient and performant compared to counterparts based on the original graph. 
\\
{\bf Graph classification:} DosCond \cite{DosCond} first extends graph condensation to graph classification tasks. DosCond utilizes a one-step gradient matching strategy to efficiently condense a large-scale graph, where the discrete structure is captured via a graph probabilistic model that can be learned in a differentiable manner. Moreover, KIDD \cite{KIDD} utilizes the kernel ridge regression to reduce the computational cost and shows strong empirical performance.

Despite the promising performance of these graph condensation methods, they highly rely on the imitation accuracy of the GNN training behavior, which limits the transferability of condensed graphs across different GNN architectures.

\section{Libraries for GNNs Acceleration}
The proliferation of GNNs has spurred the creation of graph libraries such as PyTorch Geometric (PyG) \cite{PyG} and Deep Graph Library (DGL) \cite{DGL}. However, these libraries are not optimal for large-scale graphs. Consequently, several libraries tailored for large-scale graph like EnGN \cite{EnGN}, Roc \cite{Roc}, and PSGraph \cite{PSGraph} have emerged, but they either focus on distributed scenarios or lack comprehensive support for various acceleration methods (e.g., GNN simplification methods) and diverse types of graphs (e.g., heterogeneous graphs).

\begin{figure}
    \centering
    \includegraphics[width=0.5\textwidth]{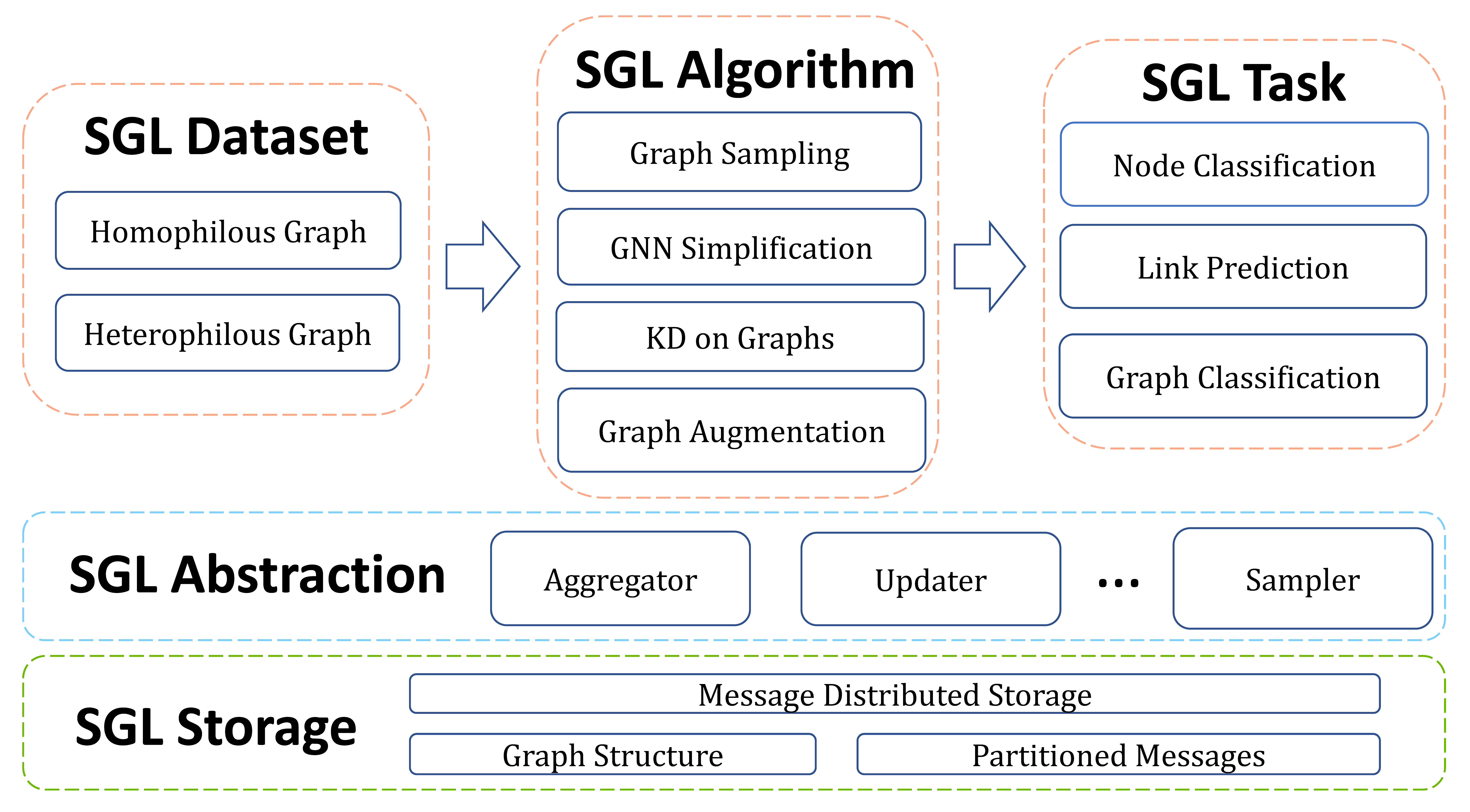}
    \caption{  
        The architecture of SGL framework
    }
    \label{fig:SGL}
\end{figure}

Addressing these issues, we present Scalable Graph Learning (SGL), a toolkit for scalable graph learning on a single machine. Figure \ref{fig:SGL} illustrates SGL's framework. The main features of SGL are as follows:
\\
{\bf  High scalability:} SGL can handle graph data with billions of nodes and edges. SGL supports both MP and SGAP paradigms for training acceleration, as well as GNN knowledge distillation for inference acceleration.
\\
{\bf  Ease of use:} SGL offers user-friendly interfaces for easily implementing and evaluating existing scalable GNNs on various downstream tasks, including node classification, node clustering, and link prediction.
\\
{\bf  Data-centric:} SGL integrates data-centric graph machine learning (DC-GML) methods (e.g., graph data augmentation methods) to enhance graph data quality and representation. Additionally, SGL also has customized acceleration algorithms for heterogeneous graphs.

\section{Future Directions}
    {\bf Complex Graph Types:} 
    Extensive research exists in graph machine learning on complex graph types, such as temporal graphs \cite{Temporal} and heterogeneous graphs \cite{Heterogeneous}. These complex types of graphs are often large-scale or rapidly expandable in practice. Consequently, it is increasingly important to design acceleration algorithms tailored for these specific types of graphs.
    \\
    {\bf Combine with DC-GML:} 
    DC-GML, including graph data collection, exploration, improvement, exploitation, maintenance, has attracted increasing attention in recent years \cite{DC-GML}. Accelerating such DC-GML methods (e.g., graph structure learning, graph feature enhancement, graph self-supervised learning, etc.) is crucial for their application to large-scale graphs.
    \\
    {\bf Customized Acceleration for Applications:}
    Customized acceleration techniques are becoming increasingly important in various applications (e.g., AI for Science \cite{chemistry,biology}, recommendation systems \cite{recommendersystems}, etc.). These applications often involve customized large-scale data and requirements, making it essential to design customized acceleration methods.

\scriptsize
\bibliographystyle{named}
\bibliography{ijcai23}

\begin{thebibliography}{}

\bibitem[\protect\citeauthoryear{Bahri \bgroup \em et al.\egroup }{2021}]{BinaryGNN}
Mehdi Bahri, Ga{\'e}tan Bahl, and Stefanos Zafeiriou.
\newblock Binary graph neural networks.
\newblock In {\em CVPR}, 2021.

\bibitem[\protect\citeauthoryear{Bai \bgroup \em et al.\egroup }{2021}]{RWT}
Jiyang Bai, Yuxiang Ren, and Jiawei Zhang.
\newblock Ripple walk training: A subgraph-based training framework for large and deep graph neural network.
\newblock In {\em IJCNN}, 2021.

\bibitem[\protect\citeauthoryear{Bing \bgroup \em et al.\egroup }{2023}]{Heterogeneous}
Rui Bing, Guan Yuan, Mu~Zhu, Fanrong Meng, Huifang Ma, and Shaojie Qiao.
\newblock Heterogeneous graph neural networks analysis: a survey of techniques, evaluations and applications.
\newblock {\em Artificial Intelligence Review}, 2023.

\bibitem[\protect\citeauthoryear{Bojchevski \bgroup \em et al.\egroup }{2020}]{PPRGo}
Aleksandar Bojchevski, Johannes Gasteiger, Bryan Perozzi, Amol Kapoor, Martin Blais, Benedek R{\'o}zemberczki, Michal Lukasik, and Stephan G{\"u}nnemann.
\newblock Scaling graph neural networks with approximate pagerank.
\newblock In {\em KDD}, 2020.

\bibitem[\protect\citeauthoryear{Chen \bgroup \em et al.\egroup }{2018a}]{VR-GCN}
Jianfei Chen, Jun Zhu, and Le~Song.
\newblock Stochastic training of graph convolutional networks with variance reduction.
\newblock {\em ICML}, 2018.

\bibitem[\protect\citeauthoryear{Chen \bgroup \em et al.\egroup }{2018b}]{FastGCN}
Jie Chen, Tengfei Ma, and Cao Xiao.
\newblock Fastgcn: fast learning with graph convolutional networks via importance sampling.
\newblock In {\em ICLR}, 2018.

\bibitem[\protect\citeauthoryear{Chen \bgroup \em et al.\egroup }{2020}]{GBP}
Ming Chen, Zhewei Wei, Bolin Ding, Yaliang Li, Ye~Yuan, Xiaoyong Du, and Ji-Rong Wen.
\newblock Scalable graph neural networks via bidirectional propagation.
\newblock In {\em NeurIPS}, 2020.

\bibitem[\protect\citeauthoryear{Chen \bgroup \em et al.\egroup }{2021}]{UGS}
Tianlong Chen, Yongduo Sui, Xuxi Chen, Aston Zhang, and Zhangyang Wang.
\newblock A unified lottery ticket hypothesis for graph neural networks.
\newblock In {\em ICML}, 2021.

\bibitem[\protect\citeauthoryear{Chen \bgroup \em et al.\egroup }{2023}]{BitGNN}
Jou-An Chen, Hsin-Hsuan Sung, Xipeng Shen, Sutanay Choudhury, and Ang Li.
\newblock Bitgnn: Unleashing the performance potential of binary graph neural networks on gpus.
\newblock In {\em ICS}, 2023.

\bibitem[\protect\citeauthoryear{Chiang \bgroup \em et al.\egroup }{2019}]{ClusterGCN}
Wei-Lin Chiang, Xuanqing Liu, Si~Si, Yang Li, Samy Bengio, and Cho-Jui Hsieh.
\newblock Cluster-gcn: An efficient algorithm for training deep and large graph convolutional networks.
\newblock In {\em KDD}, 2019.

\bibitem[\protect\citeauthoryear{Cong \bgroup \em et al.\egroup }{2020}]{MVS-GNN}
Weilin Cong, Rana Forsati, Mahmut Kandemir, and Mehrdad Mahdavi.
\newblock Minimal variance sampling with provable guarantees for fast training of graph neural networks.
\newblock In {\em KDD}, 2020.

\bibitem[\protect\citeauthoryear{Deng and Zhang}{2021}]{GFKD}
Xiang Deng and Zhongfei Zhang.
\newblock Graph-free knowledge distillation for graph neural networks.
\newblock In {\em IJCAI}, 2021.

\bibitem[\protect\citeauthoryear{Ding \bgroup \em et al.\egroup }{2021}]{VQ-GNN}
Mucong Ding, Kezhi Kong, Jingling Li, Chen Zhu, John Dickerson, Furong Huang, and Tom Goldstein.
\newblock Vq-gnn: A universal framework to scale up graph neural networks using vector quantization.
\newblock In {\em NeurIPS}, 2021.

\bibitem[\protect\citeauthoryear{Eliasof \bgroup \em et al.\egroup }{2023}]{Haar}
Moshe Eliasof, Benjamin~J Bodner, and Eran Treister.
\newblock Haar wavelet feature compression for quantized graph convolutional networks.
\newblock {\em TNNLS}, 2023.

\bibitem[\protect\citeauthoryear{Feng \bgroup \em et al.\egroup }{2020}]{SGQuant}
Boyuan Feng, Yuke Wang, Xu~Li, Shu Yang, Xueqiao Peng, and Yufei Ding.
\newblock Sgquant: Squeezing the last bit on graph neural networks with specialized quantization.
\newblock In {\em ICTAI}, 2020.

\bibitem[\protect\citeauthoryear{Feng \bgroup \em et al.\egroup }{2022}]{GRAND+}
Wenzheng Feng, Yuxiao Dong, Tinglin Huang, Ziqi Yin, Xu~Cheng, Evgeny Kharlamov, and Jie Tang.
\newblock Grand+: Scalable graph random neural networks.
\newblock In {\em WWW}, 2022.

\bibitem[\protect\citeauthoryear{Fey and Lenssen}{2019}]{PyG}
Matthias Fey and Jan~Eric Lenssen.
\newblock Fast graph representation learning with pytorch geometric.
\newblock {\em arXiv preprint arXiv:1903.02428}, 2019.

\bibitem[\protect\citeauthoryear{Fey \bgroup \em et al.\egroup }{2021}]{GNNAutoScale}
Matthias Fey, Jan~E Lenssen, Frank Weichert, and Jure Leskovec.
\newblock Gnnautoscale: Scalable and expressive graph neural networks via historical embeddings.
\newblock In {\em ICLR}, 2021.

\bibitem[\protect\citeauthoryear{Frankle and Carbin}{2019}]{LotteryTicket}
Jonathan Frankle and Michael Carbin.
\newblock The lottery ticket hypothesis: Finding sparse, trainable neural networks.
\newblock In {\em ICLR}, 2019.

\bibitem[\protect\citeauthoryear{Frasca \bgroup \em et al.\egroup }{2020}]{SIGN}
Fabrizio Frasca, Emanuele Rossi, Davide Eynard, Ben Chamberlain, Michael Bronstein, and Federico Monti.
\newblock Sign: Scalable inception graph neural networks.
\newblock In {\em ICML}, 2020.

\bibitem[\protect\citeauthoryear{Gao \bgroup \em et al.\egroup }{2018}]{LGCL}
Hongyang Gao, Zhengyang Wang, and Shuiwang Ji.
\newblock Large-scale learnable graph convolutional networks.
\newblock In {\em KDD}, 2018.

\bibitem[\protect\citeauthoryear{Gao \bgroup \em et al.\egroup }{2024}]{MCond}
Xinyi Gao, Tong Chen, Yilong Zang, Wentao Zhang, Quoc Viet~Hung Nguyen, Kai Zheng, and Hongzhi Yin.
\newblock Graph condensation for inductive node representation learning.
\newblock In {\em ICDE}, 2024.

\bibitem[\protect\citeauthoryear{Gasteiger \bgroup \em et al.\egroup }{2018}]{APPNP}
Johannes Gasteiger, Aleksandar Bojchevski, and Stephan G{\"u}nnemann.
\newblock Predict then propagate: Graph neural networks meet personalized pagerank.
\newblock In {\em ICLR}, 2018.

\bibitem[\protect\citeauthoryear{Gholami \bgroup \em et al.\egroup }{2022}]{Quantization}
Amir Gholami, Sehoon Kim, Zhen Dong, Zhewei Yao, Michael~W Mahoney, and Kurt Keutzer.
\newblock A survey of quantization methods for efficient neural network inference.
\newblock In {\em Low-Power Computer Vision}. 2022.

\bibitem[\protect\citeauthoryear{Gilmer \bgroup \em et al.\egroup }{2017}]{MPNN}
Justin Gilmer, Samuel~S Schoenholz, Patrick~F Riley, Oriol Vinyals, and George~E Dahl.
\newblock Neural message passing for quantum chemistry.
\newblock In {\em ICML}, 2017.

\bibitem[\protect\citeauthoryear{Gou \bgroup \em et al.\egroup }{2021}]{KD}
Jianping Gou, Baosheng Yu, Stephen~J Maybank, and Dacheng Tao.
\newblock Knowledge distillation: A survey.
\newblock {\em International Journal of Computer Vision}, 2021.

\bibitem[\protect\citeauthoryear{Gupta and Bedathur}{2022}]{Temporal}
Shubham Gupta and Srikanta Bedathur.
\newblock A survey on temporal graph representation learning and generative modeling.
\newblock {\em arXiv preprint arXiv:2208.12126}, 2022.

\bibitem[\protect\citeauthoryear{Hamilton \bgroup \em et al.\egroup }{2017}]{GraphSage}
Will Hamilton, Zhitao Ying, and Jure Leskovec.
\newblock Inductive representation learning on large graphs.
\newblock In {\em NeurIPS}, 2017.

\bibitem[\protect\citeauthoryear{He \bgroup \em et al.\egroup }{2022}]{KDEP}
Ruifei He, Shuyang Sun, Jihan Yang, Song Bai, and Xiaojuan Qi.
\newblock Knowledge distillation as efficient pre-training: Faster convergence, higher data-efficiency, and better transferability.
\newblock In {\em CVPR}, 2022.

\bibitem[\protect\citeauthoryear{Hu \bgroup \em et al.\egroup }{2021}]{Graph-MLP}
Yang Hu, Haoxuan You, Zhecan Wang, Zhicheng Wang, Erjin Zhou, and Yue Gao.
\newblock Graph-mlp: Node classification without message passing in graph.
\newblock {\em arXiv preprint arXiv:2106.04051}, 2021.

\bibitem[\protect\citeauthoryear{Huang \bgroup \em et al.\egroup }{2018}]{ASGCN}
Wenbing Huang, Tong Zhang, Yu~Rong, and Junzhou Huang.
\newblock Adaptive sampling towards fast graph representation learning.
\newblock In {\em NeurIPS}, 2018.

\bibitem[\protect\citeauthoryear{Huang \bgroup \em et al.\egroup }{2020}]{C&S}
Qian Huang, Horace He, Abhay Singh, Ser-Nam Lim, and Austin Benson.
\newblock Combining label propagation and simple models out-performs graph neural networks.
\newblock In {\em ICLR}, 2020.

\bibitem[\protect\citeauthoryear{Huang \bgroup \em et al.\egroup }{2022}]{EPQuant}
Linyong Huang, Zhe Zhang, Zhaoyang Du, Shuangchen Li, Hongzhong Zheng, Yuan Xie, and Nianxiong Tan.
\newblock Epquant: A graph neural network compression approach based on product quantization.
\newblock {\em NC}, 2022.

\bibitem[\protect\citeauthoryear{Huang \bgroup \em et al.\egroup }{2023}]{NIGCN}
Keke Huang, Jing Tang, Juncheng Liu, Renchi Yang, and Xiaokui Xiao.
\newblock Node-wise diffusion for scalable graph learning.
\newblock In {\em WWW}, 2023.

\bibitem[\protect\citeauthoryear{Hui \bgroup \em et al.\egroup }{2022}]{improved-UGS}
Bo~Hui, Da~Yan, Xiaolong Ma, and Wei-Shinn Ku.
\newblock Rethinking graph lottery tickets: Graph sparsity matters.
\newblock In {\em ICLR}, 2022.

\bibitem[\protect\citeauthoryear{Jia \bgroup \em et al.\egroup }{2020}]{Roc}
Zhihao Jia, Sina Lin, Mingyu Gao, Matei Zaharia, and Alex Aiken.
\newblock Improving the accuracy, scalability, and performance of graph neural networks with roc.
\newblock {\em MLSys}, 2020.

\bibitem[\protect\citeauthoryear{Jiang \bgroup \em et al.\egroup }{2020}]{PSGraph}
Jiawei Jiang, Pin Xiao, Lele Yu, Xiaosen Li, Jiefeng Cheng, Xupeng Miao, Zhipeng Zhang, and Bin Cui.
\newblock In {\em ICDE}, 2020.

\bibitem[\protect\citeauthoryear{Jin \bgroup \em et al.\egroup }{2021}]{GCOND}
Wei Jin, Lingxiao Zhao, Shichang Zhang, Yozen Liu, Jiliang Tang, and Neil Shah.
\newblock Graph condensation for graph neural networks.
\newblock In {\em ICLR}, 2021.

\bibitem[\protect\citeauthoryear{Jin \bgroup \em et al.\egroup }{2022}]{DosCond}
Wei Jin, Xianfeng Tang, Haoming Jiang, Zheng Li, Danqing Zhang, Jiliang Tang, and Bing Yin.
\newblock Condensing graphs via one-step gradient matching.
\newblock In {\em KDD}, 2022.

\bibitem[\protect\citeauthoryear{Jing \bgroup \em et al.\egroup }{2021}]{Meta-Aggregator}
Yongcheng Jing, Yiding Yang, Xinchao Wang, Mingli Song, and Dacheng Tao.
\newblock Meta-aggregator: Learning to aggregate for 1-bit graph neural networks.
\newblock In {\em ICCV}, 2021.

\bibitem[\protect\citeauthoryear{Joshi \bgroup \em et al.\egroup }{2022}]{G-CRD}
Chaitanya~K Joshi, Fayao Liu, Xu~Xun, Jie Lin, and Chuan~Sheng Foo.
\newblock On representation knowledge distillation for graph neural networks.
\newblock {\em TNNLS}, 2022.

\bibitem[\protect\citeauthoryear{Karypis and Kumar}{1997}]{METIS}
George Karypis and Vipin Kumar.
\newblock Metis: A software package for partitioning unstructured graphs, partitioning meshes, and computing fill-reducing orderings of sparse matrices.
\newblock 1997.

\bibitem[\protect\citeauthoryear{Kipf and Welling}{2017}]{GCN}
Thomas~N Kipf and Max Welling.
\newblock Semi-supervised classification with graph convolutional networks.
\newblock In {\em ICLR}, 2017.

\bibitem[\protect\citeauthoryear{Li \bgroup \em et al.\egroup }{2021}]{biology}
Rui Li, Xin Yuan, Mohsen Radfar, Peter Marendy, Wei Ni, Terence~J O'Brien, and Pablo~M Casillas-Espinosa.
\newblock Graph signal processing, graph neural network and graph learning on biological data: a systematic review.
\newblock {\em R-BME}, 2021.

\bibitem[\protect\citeauthoryear{Liang \bgroup \em et al.\egroup }{2020}]{EnGN}
Shengwen Liang, Ying Wang, Cheng Liu, Lei He, LI~Huawei, Dawen Xu, and Xiaowei Li.
\newblock Engn: A high-throughput and energy-efficient accelerator for large graph neural networks.
\newblock {\em TC}, 2020.

\bibitem[\protect\citeauthoryear{Liao \bgroup \em et al.\egroup }{2023}]{SCARA}
Ningyi Liao, Dingheng Mo, Siqiang Luo, Xiang Li, and Pengcheng Yin.
\newblock Scalable decoupling graph neural network with feature-oriented optimization.
\newblock In {\em VLDB}, 2023.

\bibitem[\protect\citeauthoryear{Liu and Zhou}{2018}]{GAT}
Zhiyuan Liu and Jie Zhou.
\newblock Graph attention networks.
\newblock In {\em ICLR}, 2018.

\bibitem[\protect\citeauthoryear{Liu \bgroup \em et al.\egroup }{2020a}]{Pruning}
Jiayi Liu, Samarth Tripathi, Unmesh Kurup, and Mohak Shah.
\newblock Pruning algorithms to accelerate convolutional neural networks for edge applications: A survey.
\newblock {\em arXiv preprint arXiv:2005.04275}, 2020.

\bibitem[\protect\citeauthoryear{Liu \bgroup \em et al.\egroup }{2020b}]{GNN-BS}
Ziqi Liu, Zhengwei Wu, Zhiqiang Zhang, Jun Zhou, Shuang Yang, Le~Song, and Qi~Yuan.
\newblock Bandit samplers for training graph neural networks.
\newblock In {\em NeurIPS}, 2020.

\bibitem[\protect\citeauthoryear{Liu \bgroup \em et al.\egroup }{2022a}]{GCDM}
Mengyang Liu, Shanchuan Li, Xinshi Chen, and Le~Song.
\newblock Graph condensation via receptive field distribution matching.
\newblock {\em arXiv preprint arXiv:2206.13697}, 2022.

\bibitem[\protect\citeauthoryear{Liu \bgroup \em et al.\egroup }{2022b}]{survey2}
Xin Liu, Mingyu Yan, Lei Deng, Guoqi Li, Xiaochun Ye, Dongrui Fan, Shirui Pan, and Yuan Xie.
\newblock Survey on graph neural network acceleration: An algorithmic perspective.
\newblock {\em arXiv preprint arXiv:2202.04822}, 2022.

\bibitem[\protect\citeauthoryear{Liu \bgroup \em et al.\egroup }{2023}]{GCEM}
Yang Liu, Deyu Bo, and Chuan Shi.
\newblock Graph condensation via eigenbasis matching.
\newblock {\em arXiv preprint arXiv:2310.09202}, 2023.

\bibitem[\protect\citeauthoryear{Page \bgroup \em et al.\egroup }{1999}]{PageRank}
Lawrence Page, Sergey Brin, Rajeev Motwani, and Terry Winograd.
\newblock The pagerank citation ranking : Bringing order to the web.
\newblock In {\em WWW}, 1999.

\bibitem[\protect\citeauthoryear{Qin \bgroup \em et al.\egroup }{2020}]{Binarization}
Haotong Qin, Ruihao Gong, Xianglong Liu, Xiao Bai, Jingkuan Song, and Nicu Sebe.
\newblock Binary neural networks: A survey.
\newblock {\em Pattern Recognition}, 2020.

\bibitem[\protect\citeauthoryear{Reiser \bgroup \em et al.\egroup }{2022}]{chemistry}
Patrick Reiser, Marlen Neubert, Andr{\'e} Eberhard, Luca Torresi, Chen Zhou, Chen Shao, Houssam Metni, Clint van Hoesel, Henrik Schopmans, Timo Sommer, et~al.
\newblock Graph neural networks for materials science and chemistry.
\newblock {\em Communications Materials}, 2022.

\bibitem[\protect\citeauthoryear{Shi \bgroup \em et al.\egroup }{2023}]{LMC}
Zhihao Shi, Xize Liang, and Jie Wang.
\newblock Lmc: Fast training of gnns via subgraph sampling with provable convergence.
\newblock In {\em ICLR}, 2023.

\bibitem[\protect\citeauthoryear{Shin and Shin}{2023}]{P&D}
Yong-Min Shin and Won-Yong Shin.
\newblock Propagate \& distill: Towards effective graph learners using propagation-embracing mlps.
\newblock In {\em LoG}, 2023.

\bibitem[\protect\citeauthoryear{Sui \bgroup \em et al.\egroup }{2021}]{Inductive}
Yongduo Sui, Xiang Wang, Tianlong Chen, Xiangnan He, and Tat-Seng Chua.
\newblock Inductive lottery ticket learning for graph neural networks.
\newblock 2021.

\bibitem[\protect\citeauthoryear{Sun \bgroup \em et al.\egroup }{2021}]{SAGN}
Chuxiong Sun, Hongming Gu, and Jie Hu.
\newblock Scalable and adaptive graph neural networks with self-label-enhanced training.
\newblock {\em arXiv preprint arXiv:2104.09376}, 2021.

\bibitem[\protect\citeauthoryear{Tailor \bgroup \em et al.\egroup }{2020}]{Degree-Quant}
Shyam~Anil Tailor, Javier Fernandez-Marques, and Nicholas~Donald Lane.
\newblock Degree-quant: Quantization-aware training for graph neural networks.
\newblock In {\em ICLR}, 2020.

\bibitem[\protect\citeauthoryear{Tian \bgroup \em et al.\egroup }{2022}]{NOSMOG}
Yijun Tian, Chuxu Zhang, Zhichun Guo, Xiangliang Zhang, and Nitesh Chawla.
\newblock Learning mlps on graphs: A unified view of effectiveness, robustness, and efficiency.
\newblock In {\em ICLR}, 2022.

\bibitem[\protect\citeauthoryear{Tian \bgroup \em et al.\egroup }{2023}]{KDG}
Yijun Tian, Shichao Pei, Xiangliang Zhang, Chuxu Zhang, and Nitesh~V Chawla.
\newblock Knowledge distillation on graphs: A survey.
\newblock {\em arXiv preprint arXiv:2302.00219}, 2023.

\bibitem[\protect\citeauthoryear{Wang \bgroup \em et al.\egroup }{2019}]{DGL}
Minjie Wang, Da~Zheng, Zihao Ye, Quan Gan, Mufei Li, Xiang Song, Jinjing Zhou, Chao Ma, Lingfan Yu, Yu~Gai, et~al.
\newblock Deep graph library: A graph-centric, highly-performant package for graph neural networks.
\newblock {\em arXiv preprint arXiv:1909.01315}, 2019.

\bibitem[\protect\citeauthoryear{Wang \bgroup \em et al.\egroup }{2021a}]{BGN}
Hanchen Wang, Defu Lian, Ying Zhang, Lu~Qin, Xiangjian He, Yiguang Lin, and Xuemin Lin.
\newblock Binarized graph neural network.
\newblock In {\em WWW}, 2021.

\bibitem[\protect\citeauthoryear{Wang \bgroup \em et al.\egroup }{2021b}]{AGP}
Hanzhi Wang, Mingguo He, Zhewei Wei, Sibo Wang, Ye~Yuan, Xiaoyong Du, and Ji-Rong Wen.
\newblock Approximate graph propagation.
\newblock In {\em KDD}, 2021.

\bibitem[\protect\citeauthoryear{Wang \bgroup \em et al.\egroup }{2021c}]{Bi-GCN}
Junfu Wang, Yunhong Wang, Zhen Yang, Liang Yang, and Yuanfang Guo.
\newblock Bi-gcn: Binary graph convolutional network.
\newblock In {\em CVPR}, 2021.

\bibitem[\protect\citeauthoryear{Wang \bgroup \em et al.\egroup }{2022}]{DLTH}
Kun Wang, Yuxuan Liang, Pengkun Wang, Xu~Wang, Pengfei Gu, Junfeng Fang, and Yang Wang.
\newblock Searching lottery tickets in graph neural networks: A dual perspective.
\newblock In {\em ICLR}, 2022.

\bibitem[\protect\citeauthoryear{Wang \bgroup \em et al.\egroup }{2023a}]{Snowflake}
Kun Wang, Guohao Li, Shilong Wang, Guibin Zhang, Kai Wang, Yang You, Xiaojiang Peng, Yuxuan Liang, and Yang Wang.
\newblock The snowflake hypothesis: Training deep gnn with one node one receptive field.
\newblock {\em arXiv preprint arXiv:2308.10051}, 2023.

\bibitem[\protect\citeauthoryear{Wang \bgroup \em et al.\egroup }{2023b}]{GC-SNTK}
Lin Wang, Wenqi Fan, Jiatong Li, Yao Ma, and Qing Li.
\newblock Fast graph condensation with structure-based neural tangent kernel.
\newblock {\em arXiv preprint arXiv:2310.11046}, 2023.

\bibitem[\protect\citeauthoryear{Wang \bgroup \em et al.\egroup }{2023c}]{QLR}
Shuang Wang, Bahaeddin Eravci, Rustam Guliyev, and Hakan Ferhatosmanoglu.
\newblock Low-bit quantization for deep graph neural networks with smoothness-aware message propagation.
\newblock In {\em CIKM}, 2023.

\bibitem[\protect\citeauthoryear{Wu \bgroup \em et al.\egroup }{2019}]{SGC}
Felix Wu, AmauriHolandade Souza, Tianyi Zhang, Christopher Fifty, Tao Yu, and KilianQ. Weinberger.
\newblock Simplifying graph convolutional networks.
\newblock In {\em ICML}, 2019.

\bibitem[\protect\citeauthoryear{Wu \bgroup \em et al.\egroup }{2022}]{recommendersystems}
Shiwen Wu, Fei Sun, Wentao Zhang, Xu~Xie, and Bin Cui.
\newblock Graph neural networks in recommender systems: a survey.
\newblock {\em ACM Computing Surveys}, 2022.

\bibitem[\protect\citeauthoryear{Wu \bgroup \em et al.\egroup }{2023}]{KRD}
Lirong Wu, Haitao Lin, Yufei Huang, and Stan~Z Li.
\newblock Quantifying the knowledge in gnns for reliable distillation into mlps.
\newblock In {\em ICML}, 2023.

\bibitem[\protect\citeauthoryear{Xu \bgroup \em et al.\egroup }{2023}]{KIDD}
Zhe Xu, Yuzhong Chen, Menghai Pan, Huiyuan Chen, Mahashweta Das, Hao Yang, and Hanghang Tong.
\newblock Kernel ridge regression-based graph dataset distillation.
\newblock In {\em KDD}, 2023.

\bibitem[\protect\citeauthoryear{Yan \bgroup \em et al.\egroup }{2020}]{TinyGNN}
Bencheng Yan, Chaokun Wang, Gaoyang Guo, and Yunkai Lou.
\newblock Tinygnn: Learning efficient graph neural networks.
\newblock In {\em KDD}, 2020.

\bibitem[\protect\citeauthoryear{Yang \bgroup \em et al.\egroup }{2020}]{LSP}
Yiding Yang, Jiayan Qiu, Mingli Song, Dacheng Tao, and Xinchao Wang.
\newblock Distilling knowledge from graph convolutional networks.
\newblock In {\em CVPR}, 2020.

\bibitem[\protect\citeauthoryear{Yang \bgroup \em et al.\egroup }{2021}]{CPF}
Cheng Yang, Jiawei Liu, and Chuan Shi.
\newblock Extract the knowledge of graph neural networks and go beyond it: An effective knowledge distillation framework.
\newblock In {\em WWW}, 2021.

\bibitem[\protect\citeauthoryear{Yang \bgroup \em et al.\egroup }{2022}]{GKD}
Chenxiao Yang, Qitian Wu, and Junchi Yan.
\newblock Geometric knowledge distillation: Topology compression for graph neural networks.
\newblock In {\em NeurIPS}, 2022.

\bibitem[\protect\citeauthoryear{Yang \bgroup \em et al.\egroup }{2024}]{VQGraph}
Ling Yang, Ye~Tian, Minkai Xu, Zhongyi Liu, Shenda Hong, Wei Qu, Wentao Zhang, Bin Cui, Muhan Zhang, and Jure Leskovec.
\newblock Vqgraph: Graph vector-quantization for bridging gnns and mlps.
\newblock In {\em ICLR}, 2024.

\bibitem[\protect\citeauthoryear{Yao and Li}{2021}]{BNS}
Kai-Lang Yao and Wu-Jun Li.
\newblock Blocking-based neighbor sampling for large-scale graph neural networks.
\newblock In {\em IJCAI}, 2021.

\bibitem[\protect\citeauthoryear{Ying \bgroup \em et al.\egroup }{2018}]{PinSage}
Rex Ying, Ruining He, Kaifeng Chen, Pong Eksombatchai, William~L. Hamilton, and Jure Leskovec.
\newblock Graph convolutional neural networks for web-scale recommender systems.
\newblock In {\em KDD}, 2018.

\bibitem[\protect\citeauthoryear{Yoon \bgroup \em et al.\egroup }{2021}]{PASS}
Minji Yoon, Th{\'e}ophile Gervet, Baoxu Shi, Sufeng Niu, Qi~He, and Jaewon Yang.
\newblock Performance-adaptive sampling strategy towards fast and accurate graph neural networks.
\newblock In {\em KDD}, 2021.

\bibitem[\protect\citeauthoryear{You \bgroup \em et al.\egroup }{2022}]{GEB}
Haoran You, Zhihan Lu, Zijian Zhou, Yonggan Fu, and Yingyan Lin.
\newblock Early-bird gcns: Graph-network co-optimization towards more efficient gcn training and inference via drawing early-bird lottery tickets.
\newblock In {\em AAAI}, 2022.

\bibitem[\protect\citeauthoryear{Younesian \bgroup \em et al.\egroup }{2023}]{GRAPES}
Taraneh Younesian, Thiviyan Thanapalasingam, Emile van Krieken, Daniel Daza, and Peter Bloem.
\newblock Grapes: Learning to sample graphs for scalable graph neural networks.
\newblock In {\em NeurIPS}, 2023.

\bibitem[\protect\citeauthoryear{Zeng \bgroup \em et al.\egroup }{2019}]{GraphSAINT}
Hanqing Zeng, Hongkuan Zhou, Ajitesh Srivastava, Rajgopal Kannan, and Viktor Prasanna.
\newblock Graphsaint: Graph sampling based inductive learning method.
\newblock In {\em ICLR}, 2019.

\bibitem[\protect\citeauthoryear{Zeng \bgroup \em et al.\egroup }{2021}]{SHADOW-GNN}
Hanqing Zeng, Muhan Zhang, Yinglong Xia, Ajitesh Srivastava, Andrey Malevich, Rajgopal Kannan, Viktor Prasanna, Long Jin, and Ren Chen.
\newblock Decoupling the depth and scope of graph neural networks.
\newblock In {\em NeurIPS}, 2021.

\bibitem[\protect\citeauthoryear{Zhang \bgroup \em et al.\egroup }{2021a}]{GLNN}
Shichang Zhang, Yozen Liu, Yizhou Sun, and Neil Shah.
\newblock Graph-less neural networks: Teaching old mlps new tricks via distillation.
\newblock In {\em ICLR}, 2021.

\bibitem[\protect\citeauthoryear{Zhang \bgroup \em et al.\egroup }{2021b}]{NDLS}
Wentao Zhang, Mingyu Yang, Zeang Sheng, Yang Li, Wen Ouyang, Yangyu Tao, Zhi Yang, and Bin Cui.
\newblock Node dependent local smoothing for scalable graph learning.
\newblock In {\em NeurIPS}, 2021.

\bibitem[\protect\citeauthoryear{Zhang \bgroup \em et al.\egroup }{2022a}]{pAScA}
Wentao Zhang, Yu~Shen, Zheyu Lin, Yang Li, Xiaosen Li, Wen Ouyang, Yangyu Tao, Zhi Yang, and Bin Cui.
\newblock Pasca: A graph neural architecture search system under the scalable paradigm.
\newblock In {\em WWW}, 2022.

\bibitem[\protect\citeauthoryear{Zhang \bgroup \em et al.\egroup }{2022b}]{NAFS}
Wentao Zhang, Zeang Sheng, Mingyu Yang, Yang Li, Yu~Shen, Zhi Yang, and Bin Cui.
\newblock Nafs: A simple yet tough-to-beat baseline for graph representation learning.
\newblock In {\em ICML}, 2022.

\bibitem[\protect\citeauthoryear{Zhang \bgroup \em et al.\egroup }{2022c}]{GAMLP}
Wentao Zhang, Ziqi Yin, Zeang Sheng, Yang Li, Wen Ouyang, Xiaosen Li, Yangyu Tao, Zhi Yang, and Bin Cui.
\newblock Graph attention multi-layer perceptron.
\newblock In {\em KDD}, 2022.

\bibitem[\protect\citeauthoryear{Zhang \bgroup \em et al.\egroup }{2022d}]{ANS-GT}
Zaixi Zhang, Qi~Liu, Qingyong Hu, and Chee-Kong Lee.
\newblock Hierarchical graph transformer with adaptive node sampling.
\newblock In {\em NeurIPS}, 2022.

\bibitem[\protect\citeauthoryear{Zhang \bgroup \em et al.\egroup }{2023}]{survey1}
Shichang Zhang, Atefeh Sohrabizadeh, Cheng Wan, Zijie Huang, Ziniu Hu, Yewen Wang, Jason Cong, Yizhou Sun, et~al.
\newblock A survey on graph neural network acceleration: Algorithms, systems, and customized hardware.
\newblock {\em arXiv preprint arXiv:2306.14052}, 2023.

\bibitem[\protect\citeauthoryear{Zhao \bgroup \em et al.\egroup }{2020}]{LPGNAS}
Yiren Zhao, Duo Wang, Daniel Bates, Robert Mullins, Mateja Jamnik, and Pietro Lio.
\newblock Learned low precision graph neural networks.
\newblock {\em arXiv preprint arXiv:2009.09232}, 2020.

\bibitem[\protect\citeauthoryear{Zheng \bgroup \em et al.\egroup }{2023a}]{DC-GML}
Xin Zheng, Yixin Liu, Zhifeng Bao, Meng Fang, Xia Hu, Alan Wee-Chung Liew, and Shirui Pan.
\newblock Towards data-centric graph machine learning: Review and outlook.
\newblock {\em arXiv preprint arXiv:2309.10979}, 2023.

\bibitem[\protect\citeauthoryear{Zheng \bgroup \em et al.\egroup }{2023b}]{SFGC}
Xin Zheng, Miao Zhang, Chunyang Chen, Quoc Viet~Hung Nguyen, Xingquan Zhu, and Shirui Pan.
\newblock Structure-free graph condensation: From large-scale graphs to condensed graph-free data.
\newblock {\em arXiv preprint arXiv:2306.02664}, 2023.

\bibitem[\protect\citeauthoryear{Zhou \bgroup \em et al.\egroup }{2021}]{Channel-Pruning}
Hongkuan Zhou, Ajitesh Srivastava, Hanqing Zeng, Rajgopal Kannan, and ViktorK. Prasanna.
\newblock Accelerating large scale real-time gnn inference using channel pruning.
\newblock In {\em VLDB}, 2021.

\bibitem[\protect\citeauthoryear{Zhu and Koniusz}{2021}]{S2GC}
Hao Zhu and Piotr Koniusz.
\newblock Simple spectral graph convolution.
\newblock In {\em ICLR}, 2021.

\bibitem[\protect\citeauthoryear{Zhu \bgroup \em et al.\egroup }{2022}]{A2Q}
Zeyu Zhu, Fanrong Li, Zitao Mo, Qinghao Hu, Gang Li, Zejian Liu, Xiaoyao Liang, and Jian Cheng.
\newblock $\mathrm{A}^2\mathrm{Q}$: Aggregation-aware quantization for graph neural networks.
\newblock In {\em ICLR}, 2022.

\bibitem[\protect\citeauthoryear{Zou \bgroup \em et al.\egroup }{2019}]{LADIES}
Difan Zou, Ziniu Hu, Yewen Wang, Song Jiang, Yizhou Sun, and Quanquan Gu.
\newblock Layer-dependent importance sampling for training deep and large graph convolutional networks.
\newblock In {\em NeurIPS}, 2019.

\end{thebibliography}

\end{document}